\documentclass[journal=jacsat,manuscript=article]{achemso}

\usepackage[version=3]{mhchem} % Formula subscripts using \ce{}

\usepackage{array}
\usepackage{amsmath}
\usepackage{amsfonts}
\usepackage{graphicx}
\usepackage{bm}
\usepackage{bbm}
\usepackage{booktabs}
\usepackage{xcolor} % for colors
\usepackage{booktabs, tabularx, threeparttable} % for professional tables
\usepackage{xspace}
\usepackage{hyperref}
\usepackage{url}
\usepackage{cleveref}
\usepackage{dsfont}
\usepackage{textalpha}
\usepackage{caption}

\mciteErrorOnUnknownfalse

\author{Vincent Fan}
\affiliation{Computer Science and Artificial Intelligence Laboratory, Massachusetts Institute of Technology, Cambridge, MA, 02139}
\altaffiliation{Equal Contribution}
\email{vincentf@mit.edu}
\author{Yujie Qian}
\affiliation{Computer Science and Artificial Intelligence Laboratory, Massachusetts Institute of Technology, Cambridge, MA, 02139}
\altaffiliation{Equal Contribution}
\email{yujieq@csail.mit.edu}
\author{Alex Wang}
\affiliation{Computer Science and Artificial Intelligence Laboratory, Massachusetts Institute of Technology, Cambridge, MA, 02139}
\altaffiliation{Equal Contribution}
\author{Amber Wang}
\affiliation{Computer Science and Artificial Intelligence Laboratory, Massachusetts Institute of Technology, Cambridge, MA, 02139}
\author{Connor W. Coley}
\affiliation{Department of Chemical Engineering, Massachusetts Institute of Technology, Cambridge, MA, 02139}
\alsoaffiliation{Department of Electrical Engineering and Computer Science, Massachusetts Institute of Technology, Cambridge, MA, 02139}
\author{Regina Barzilay}
\email{regina@csail.mit.edu}
\affiliation{Computer Science and Artificial Intelligence Laboratory, Massachusetts Institute of Technology, Cambridge, MA, 02139}
\alsoaffiliation{Department of Electrical Engineering and Computer Science, Massachusetts Institute of Technology, Cambridge, MA, 02139}

\title{OpenChemIE: An Information Extraction Toolkit For Chemistry Literature}

\newcommand{\ours}{\text{OpenChemIE}\xspace}

\begin{document}

\begin{abstract}
Information extraction from chemistry literature is vital for constructing up-to-date reaction databases for data-driven chemistry. Complete extraction requires combining information across text, tables, and figures, whereas prior work has mainly investigated extracting reactions from single modalities. In this paper, we present \ours to address this complex challenge and enable the extraction of reaction data at the document level. \ours approaches the problem in two steps: extracting relevant information from individual modalities and then integrating the results to obtain a final list of reactions. For the first step, we employ specialized neural models that each address a specific task for chemistry information extraction, such as parsing molecules or reactions from text or figures. We then integrate the information from these modules using chemistry-informed algorithms, allowing for the extraction of fine-grained reaction data from reaction condition and substrate scope investigations. Our machine learning models attain state-of-the-art performance when evaluated individually, and we meticulously annotate a challenging dataset of reaction schemes with R-groups to evaluate our pipeline as a whole, achieving an F1 score of 69.5\%. Additionally, the reaction extraction results of \ours attain an accuracy score of 64.3\% when directly compared against the Reaxys chemical database. We provide \ours freely to the public as an open-source package, as well as through a web interface.

\end{abstract}

\section{Introduction}

Reaction data curated from scientific literature is commonly used to train models for cheminformatics. Today, this data is collected and maintained by experts in databases such as Reaxys \cite{reaxys}. However, this manual extraction comes with prohibitive cost and delayed updates. Moreover, increasingly nuanced machine learning models for reaction development require more fine-grained and comprehensive data, pertaining to reaction conditions, substrate scope, and other screening processes in synthetic chemistry\cite{D2SC05089G, maser, gaorxn, textreact}. Existing automated techniques can only partially address this task, focusing on specific subproblems, such as reaction parsing from individual diagrams or text passages\cite{RxnScribe, chemrxnextractor, CSR, chemdataextractor}. In this paper, we present \ours, a system that extracts reaction data from chemical literature at the document level. 

\begin{figure}[t!]
    \centering
    \includegraphics[width=0.84\linewidth]{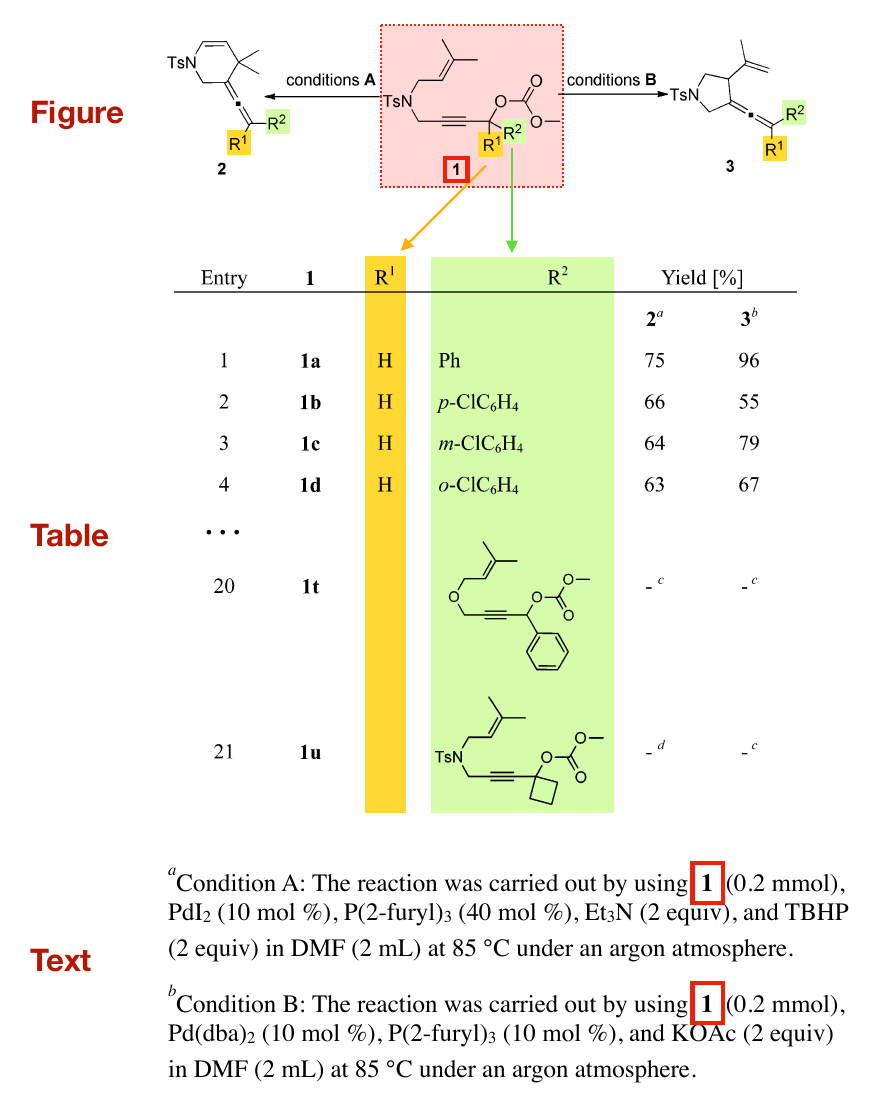}
    \caption{
    Example of a multimodal reaction description drawn from \citeauthor{doi:10.1021/jo202590w}\cite{doi:10.1021/jo202590w}. The reaction template is displayed in a figure, but information regarding R-groups is only contained in the highlighted sections of the table. Moreover, detailed reaction conditions are described in the table and accompanying footnotes.}
    \label{fig:lit example}
\end{figure}

This extraction task is difficult because large swathes of reaction data are realized in multiple modalities, often requiring chemical reasoning to fully determine relevant molecular structures. \Cref{fig:lit example} illustrates two challenges. First, the molecular structures are not entirely depicted in the figure, as they contain R-groups. The abbreviated structures can be identified by directly parsing phenol for R$^2$ in entry 1 of the accompanying table, or by comparing the differences in the molecular graphs of \textbf{1} and \textbf{1u} in entry 21. Second, the system must align additional reaction metadata with the correct structures. In \Cref{fig:lit example} the highlighted molecule is only referred to by the label \textbf{1} in the footnote text, which contains detailed reaction conditions. In other cases, conditions may also be defined in a figure or table, or the reaction itself may be described in text. 

To address these challenges, \ours provides a streamlined computational pipeline, which analyzes individual modalities and combines the extracted information together to recover implicitly defined reactions. Building on our prior research in reaction extraction\cite{chemrxnextractor}, molecular optical recognition\cite{MolScribe}, and reaction diagram parsing\cite{RxnScribe}, we design additional modules that enable \ours to fuse information at three different levels. First, we train a machine learning model to associate molecules depicted in diagrams with their text labels, performing a multimodal coreference resolution. Second, \ours aligns reactions with reaction conditions and other data presented in tables, annotated in figures, or discussed in texts by utilizing the coreference information. Lastly, \ours recognizes R-groups in a diagram by comparing molecules with the same label, and substitutes them with additional substructures listed in substrate scope tables and figures, yielding complete substrate data. 

We evaluate the performance of each individual machine learning module as well as the system as a whole. To evaluate the overall performance of \ours, we manually curated a dataset of 1007 reactions described in 78 substrate scope figures involving R-groups across five different organic chemistry journals. The extraction task requires all reaction components to be correctly predicted and R-groups to be resolved. \ours achieves an F1 score of 69.5\% on this dataset, and we performed a meticulous evaluation to analyze the error contributions of the different modules of the pipeline, identifying areas where the system could be further enhanced in future work. Furthermore, in an end-to-end evaluation of \ours on extracting reaction data from journal articles, we attain an accuracy of 64.3\% when comparing against existing extractions in Reaxys. Notably, our models for reaction diagram parsing, molecule detection, and coreference resolution all perform robustly under evaluation on independent benchmarks, and our R-group resolution algorithm only contributes to a small amount of mistakes. The majority of errors were due to mistakes in molecule recognition or optical character recognition.

\ours is available on a public web portal (\url{https://mit.openchemie.info}) as an easily accessible demonstration of key forms of analysis that we incorporate. The full pipeline and its individual methods for analysis are provided in a Python package (\url{https://github.com/CrystalEye42/OpenChemIE}) that is suitable for larger-scale information extraction. Our Python package allows for comprehensive extractions of molecules and reactions from PDF files, as well as from only text and images. The toolkit is fully open-source to facilitate future development in this area.  
\section{Related Work} 

\paragraph{Extracting From Figures} This task includes molecule recognition and reaction extraction. Molecule recognition involves translating molecular images into SMILES strings. Initial approaches employed rule-based methods for determining the structures of molecules, utilizing a suite of algorithms and heuristics to detect individual components such as bonds and atoms \cite{osra, molvec}. Later works instead leveraged CNN-based encoder-decoder architectures from deep learning to perform this segmentation, allowing for robust recognition across diverse styles \cite{mse-staker,rajan2020review,chemgrapher,img2mol,rajan2021decimer,MolScribe}. Recent research has also enabled the extraction of reaction schemes from figures, in which the reactants, products, conditions, and yield for each reaction are identified. These works approached the segmentation either by using a series of heuristics and image filters\cite{reactiondataextractor} or by applying data-driven models for object detection and sequence generation\cite{RxnScribe,reactiondataextractor2}. 

Several works have additionally developed systems for automatic extraction from figures of real-world documents. These include ReactionDataExtractor \cite{reactiondataextractor} and its 2.0 version \cite{reactiondataextractor2}, which involve parsing reaction schemes, and also include MolMiner \cite{molminer} and DECIMER.ai \cite{decimer.ai}, which focus on molecule recognition. ChemSchematicResolver\cite{CSR} is another molecule recognition system that additionally resolves R-groups defined in text labels. However, achieving a robust figure-based extraction system remains challenging due to the wide variation of styles and possible complexities for molecules and reaction schemes.

\paragraph{Extracting From Text} This task consists of identifying chemical entities and their roles, as well as parsing described reactions. Several studies for the former have centered on dataset curation\cite{CHEMDNER,NguyenZYFDTHACB20}. For both chemical entity identification and reaction extraction, proposed solutions include parsers that employ a series of regular expressions and classifiers to detect key terms\cite{JessopAWHM11,chemicaltagger}. A deep-learning solution for extracting reactions instead formulates the problem as a sequence labeling task and utilizes a fine-tuned transformer encoder architecture. 

A few works have created systems that extract from PDF files of documents instead of from plain text, presenting additional engineering challenges. These include ChemDataExtractor\cite{chemdataextractor} and PDFDataExtractor\cite{pdfdataextractor}, which identify chemical entities and their associated properties. While these can extract important information available in text, they do not process relevant information from figures that would augment the data. Text-based descriptions of reactions and molecules are often underspecified, generally referring to these entities using families of compounds or by labels defined in figures. Resolving these mentions to obtain specific molecular structures is vital. In contrast to both these and the figure-based extraction systems, \ours aims for a more versatile, unified system. The advantage of \ours is our usage of specialized chemistry-informed algorithms to integrate extractions from multiple modalities, namely text, tables, and figures, thus overcoming the single modality barrier and enabling more comprehensive extractions. 
\section{Problem Formulation}

\begin{figure}[t]
    \centering
    \includegraphics[width=\linewidth]{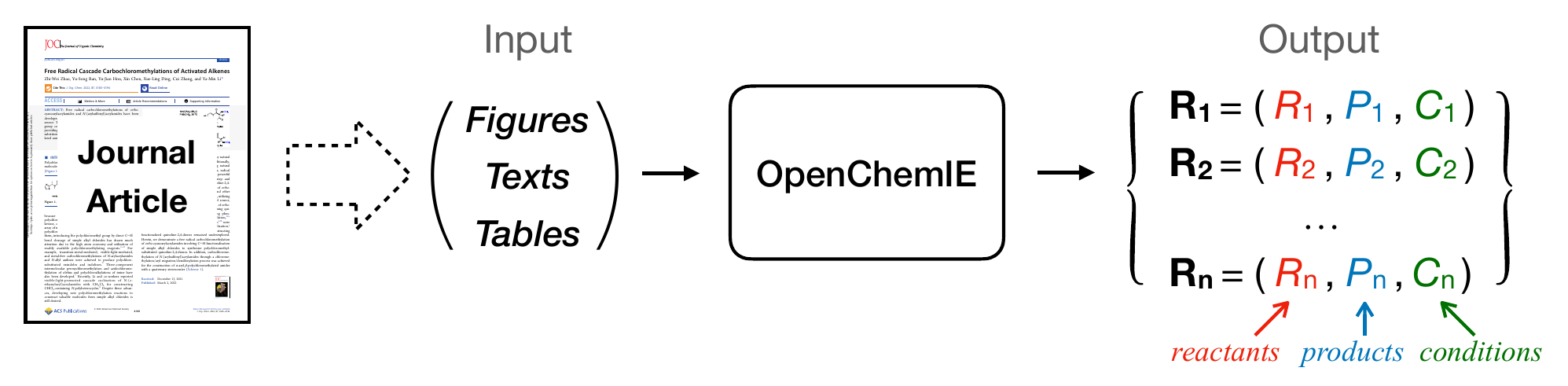}
    \caption{
    \ours addresses the problem of extracting a list of reactions, containing chemical structures for reactants and products, as well as reaction conditions from a PDF document segmented into figures, text, and tables.}
    \label{fig:formulation}
\end{figure}

As seen in \Cref{fig:formulation}, \ours addresses the task of extracting detailed chemical reactions at the document level. We consider the input journal article to be a triple of (Figures, Texts, Tables), which can be automatically segmented with existing PDF parsing tools. We seek to extract the reactions described in the paper by identifying machine-readable structures for their reactants and products, as well as other metadata. The expected output is a list of reactions $\{\mathbf{R_1},\mathbf{R_2},\dots, \mathbf{R_n}\}$, where each reaction $\mathbf{R_i}$ is a triple $(R_i, P_i, C_i)$. $R_i$ is the set of reactants and $P_i$ is the set of products, each consisting of one or more molecules expressed as SMILES strings. $C_i$ is the set of metadata associated with the reaction, including detailed conditions and yield information, and may be empty if no such information is parsed from the paper. We do not capture information contained in other plots, such as reaction coordinate diagrams or spectral data.

The information extraction task is thus expressed as a function 
\begin{equation}
 f : (\text{Figures, Texts, Tables}) \rightarrow \{\mathbf{R}_1, \mathbf{R}_2, \dots, \mathbf{R}_n\}
\end{equation}
Crucially, \ours establishes relationships between its three inputs to inform its output, such that $f(\text{Figures, Text, Tables})$ contains more data than $f(\text{Figures})\cup f(\text{Texts}) \cup f(\text{Tables})$, the result of individually extracting from each modality. 

\section{\ours Overview}

\begin{figure}[t]
    \centering
    \includegraphics[width=\linewidth]{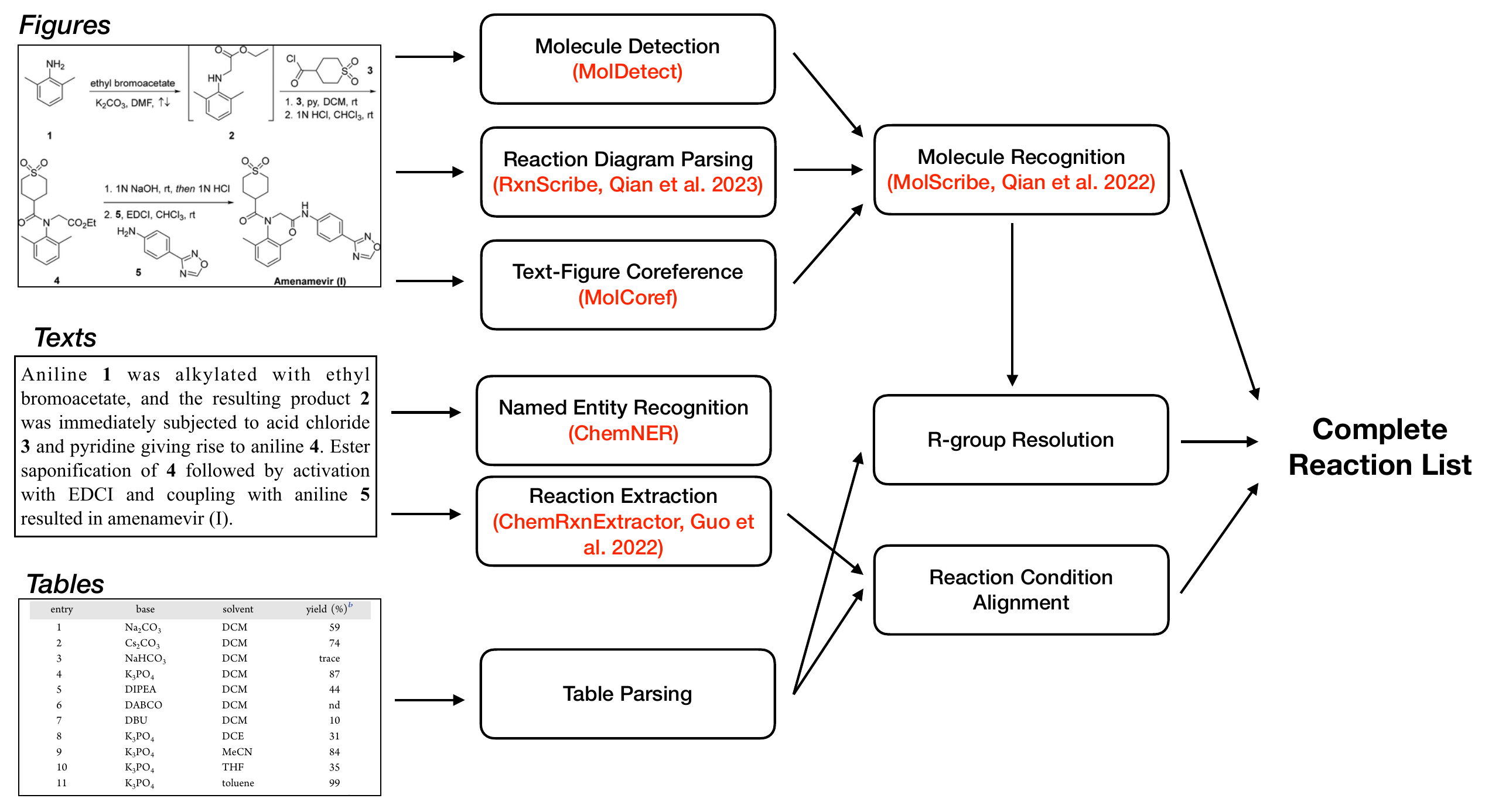}
    \caption{Overview of \ours, which receives segmented figures, texts,and tables for processing. The results from individual neural models for each modality are combined through reaction condition alignment and R-group resolution in \ours to yield a final list of reactions.}
    \label{fig:overview}
\end{figure}

In the following section, we present an overview of \ours, a dedicated toolkit designed to extract full reaction data from chemistry papers. A summary of our system can be found in \Cref{fig:overview}. Initially, \ours receives the document which has been segmented into figures, text, and tables for use in the downstream steps of our pipeline (implementation details are provided in the Supporting Information). For each modality, we have developed specialized machine learning models capable of effectively parsing molecules and reactions, as well as inferring relationships between text and diagrams. To further expand the scope of information captured by \ours, we implement two general procedures that fuse the outputs of individual models to produce a more complete reaction list. A description of the individual components of \ours follows. 

\begin{itemize}
    \item \textbf{Figure Analysis.} Analysis of chemistry figures requires strong visual understanding, ranging from high-level comprehension of reaction schemes and relations between entities to low-level recognition of molecules. To address this multifaceted challenge, we provide four models for figure/scheme analysis. The first of these models is designed for detecting sub-images of molecules within figures (molecule detection, MolDetect) and providing the relevant bounding box. Another model is for resolving the coreference between detected molecules in the figure and labels in the text (text-figure coreference, MolCoref). Additionally, \ours utilizes our previous research for parsing reaction schemes and relevant condition information (reaction diagram parsing, RxnScribe) \cite{RxnScribe} and translating molecular images into their chemical structures (molecule recognition, MolScribe) \cite{MolScribe}.
    \item \textbf{Text Analysis.} Extracting from chemistry texts involves identifying mentions of molecules and chemical reactions. To this end, we provide two models to address both of these subtasks. The first one is a model for extracting chemical entities from texts (named entity recognition, ChemNER). The second model, from our previous research, identifies chemical reactions and their reaction conditions (reaction extraction, ChemRxnExtractor)\cite{chemrxnextractor}. 
    \item \textbf{Multimodal Integration.} We implement two additional procedures that integrate information across our single modality models. R-group resolution is the process of identifying and substituting R-group structures into reaction templates, which allows for a more complete extraction of reactions. This process utilizes parsed data from tables as well as molecule information from figure analysis models. Reaction condition alignment enhances text-based reaction descriptions with molecular structures identified in relevant diagrams. These multimodal integration components enable \ours to serve as an end-to-end pipeline for reaction extraction. 
    
\end{itemize}

In the subsequent sections, we will elaborate on the technical details of each module comprising \ours. 

\section{Figure Analysis}

\begin{figure}[t]
    \centering
    \includegraphics[width=\linewidth]{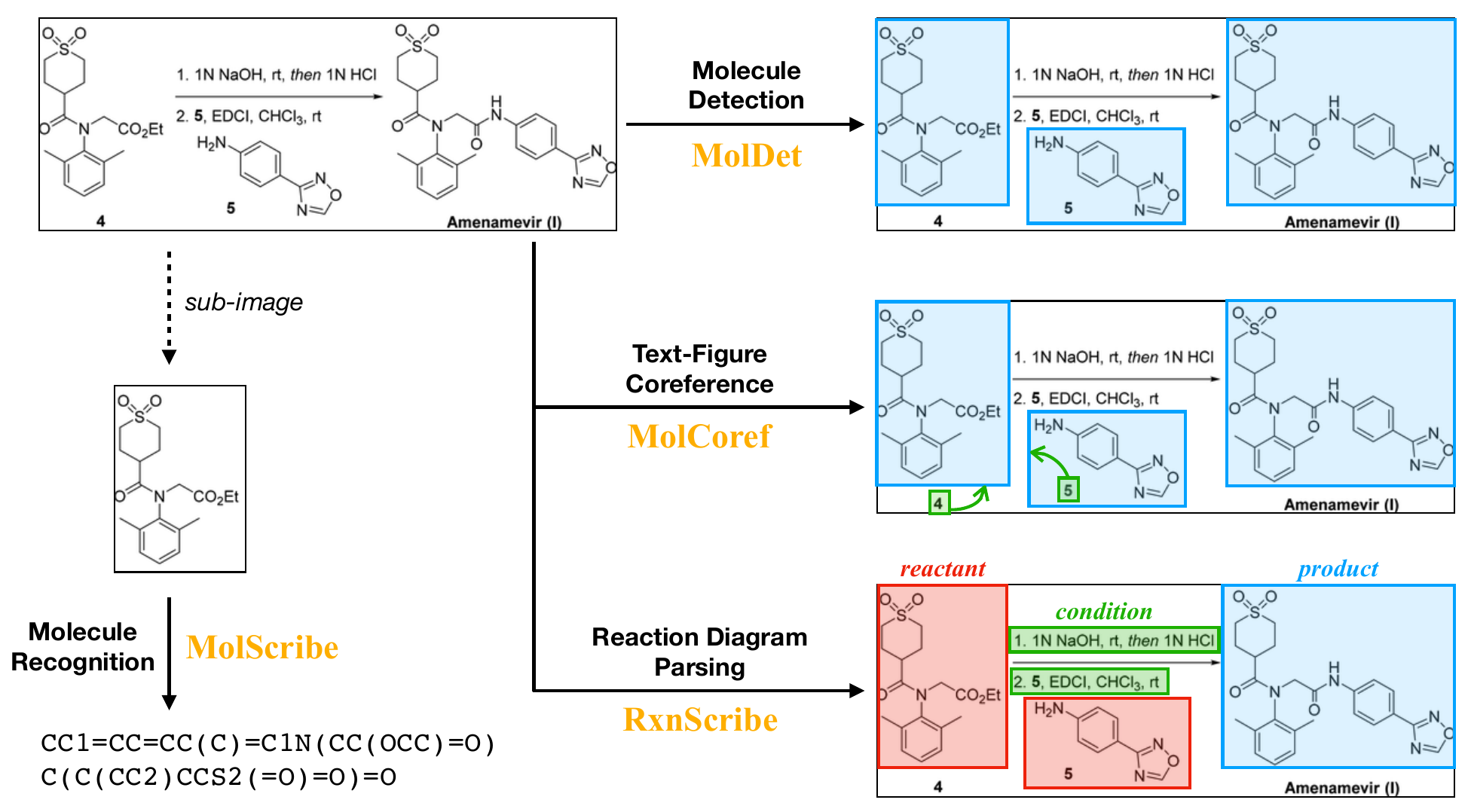}
    \caption{OpenChemIE provides four models for analyzing figures in chemistry literature, including molecule detection, text-figure coreference, reaction diagram parsing, and molecule recognition.}
    \label{fig:figuretasks}
\end{figure}

Figures in chemistry literature contain essential molecular structures and reactions, as well as relational information to text in or surrounding the diagram. In particular, the new methods we develop focus on this latter aspect, which enable the success of our downstream multimodal integration modules. As in \Cref{fig:figuretasks}, \ours addresses four facets of figure analysis, which we detail in this section. 

\paragraph{Molecule Detection}

In chemistry literature, a single figure often contains multiple molecules. The task of molecule detection is to segment the figure into sub-images of molecules such that we can later recognize the structure of each individual molecule. Molecule detection shares similarities with the extensively studied object detection task in computer vision \cite{DETR,Pix2Seq}, which focuses on identifying sub-images of objects within natural photographs.

In \ours, we provide MolDetect, a molecule detection model formulated with sequence generation. Inspired by the Pix2Seq \cite{Pix2Seq} model designed for object detection, MolDetect identifies molecular sub-images by predicting their bounding boxes as a sequence. Given a figure, a molecule entity whose bounding box has top-left coordinates $(\mathrm{x_1}, \mathrm{y_1})$ and bottom-right coordinates $(\mathrm{x_2}, \mathrm{y_2})$ is represented as five discrete tokens,
\begin{equation}
    \text{Molecule} := \ \mathrm{x_1} \ \mathrm{y_1} \ \mathrm{x_2} \ \mathrm{y_2} \ \texttt{[Mol]}
    \label{eq:entity}
\end{equation}
where \texttt{[Mol]} is a special token indicating the detection of a molecule. MolDetect sequentially generates all the molecule entities within the figure, 
\begin{equation}
    \text{MolDetectOutput} := (\text{Molecule})^*
    \label{eq:moldet}
\end{equation}
where $(\cdot)^*$ means zero or more occurrences. 

MolDetect is implemented as an encoder-decoder architecture. The figure is encoded using a convolutional neural network to obtain hidden representations. Then, the decoder is a Transformer which generates the output sequence as defined in \Cref{eq:entity,eq:moldet}.

\paragraph{Text-Figure Coreference}

\begin{figure}[t]
    \centering
    \includegraphics[width = 0.8\linewidth]{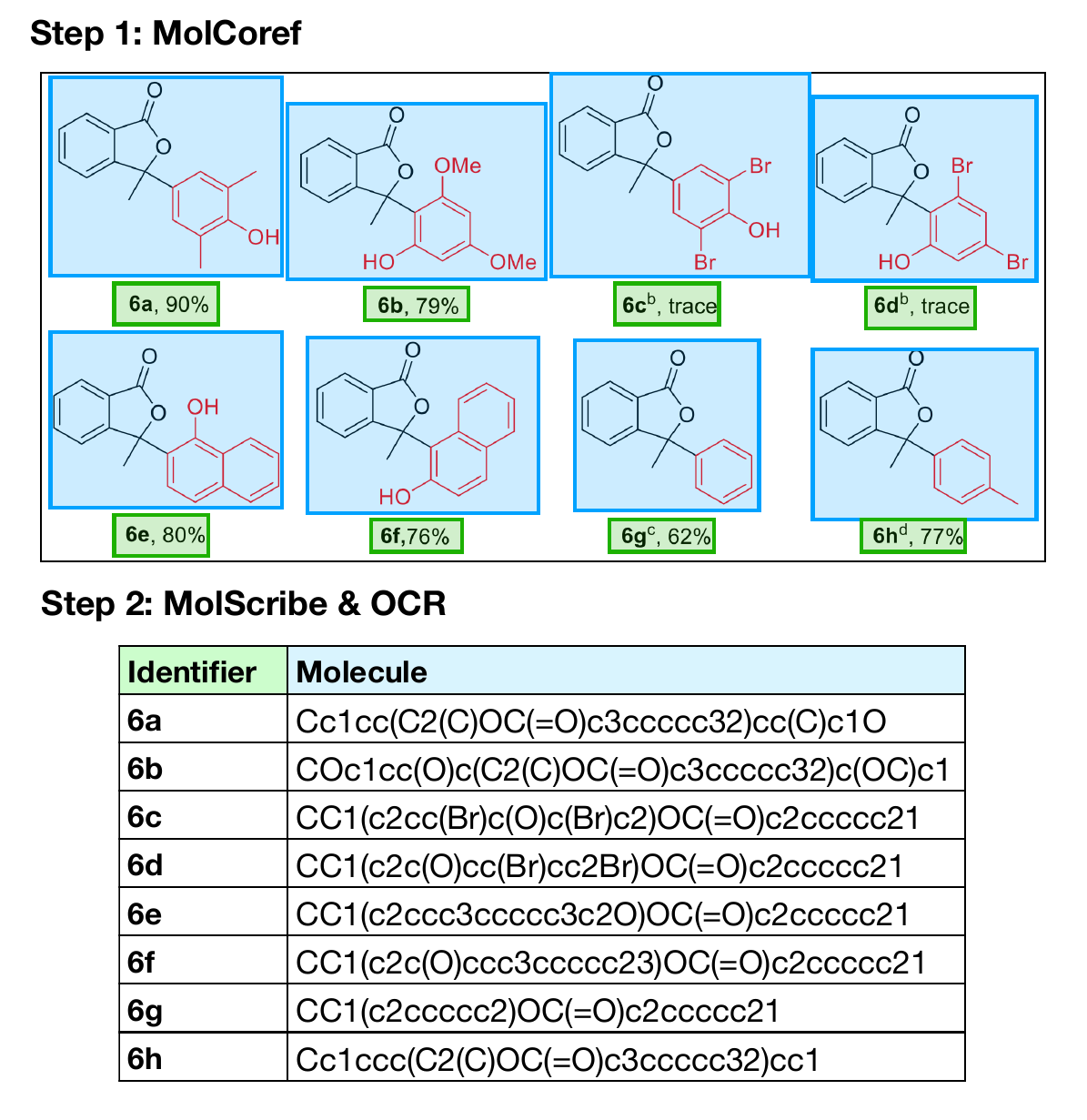}
    \caption{Illustration of extracting molecules and identifier coreferences from a figure. First, MolCoref determines entity bounding boxes and their correspondence. Then, the molecules are recognized by MolScribe.}
    \label{fig:mol_from_figure}
\end{figure}

It is common practice in chemistry literature to assign unique identifiers to the molecules depicted in a figure and subsequently refer to them by their respective identifiers in the accompanying text. To establish a clear link between the information in the text and the figure, we have developed MolCoref, a model that pairs molecules with their respective identifiers in the figure. An example prediction from MolCoref can be found in \Cref{fig:mol_from_figure}, which depicts the molecules and identifiers being successfully detected and the corresponding links established.

MolCoref also employs a sequence generation approach to resolve the coreference between identifiers and molecular structures. Specifically, its output format is defined as 
\begin{equation}
\begin{split}
    & \text{MolCorefOutput} := \ ( \text{Molecule}\ [\text{Identifier}]? )^* \\
    & \text{Molecule} := \ \mathrm{x_1} \ \mathrm{y_1} \ \mathrm{x_2} \ \mathrm{y_2} \ \texttt{[Mol]} \\
    & \text{Identifier} := \ \mathrm{x_1} \ \mathrm{y_1} \ \mathrm{x_2} \ \mathrm{y_2} \ \texttt{[Idt]}
\end{split}
\end{equation}
where $[\cdot]?$ means optional. Both the Molecule and Identifier are represented using five tokens, consisting of four coordinates and a final token to differentiate them. When a Molecule is paired with an Identifier in the figure, the model generates the Molecule first, followed by the corresponding Identifier. Otherwise, the model generates only the Molecule without the Identifier. Based on MolCoref's output, we use a molecular recognition model to parse the chemical structures and an optical character recognition (OCR) model to parse the text strings.

Compared to the previous approach \cite{CSR} that relies on heuristic rules for aligning molecules with their identifiers, MolCoref integrates the detection of molecule bounding boxes and the resolution of text-figure coreference into a single model. This simplifies the process and mitigates the risk of error propagation. Furthermore, as our experiments will demonstrate, our data-driven model yields more accurate and reliable predictions.

\paragraph{Reaction Diagram Parsing}

\begin{figure}[t]
    \centering
    \includegraphics[width=\linewidth]{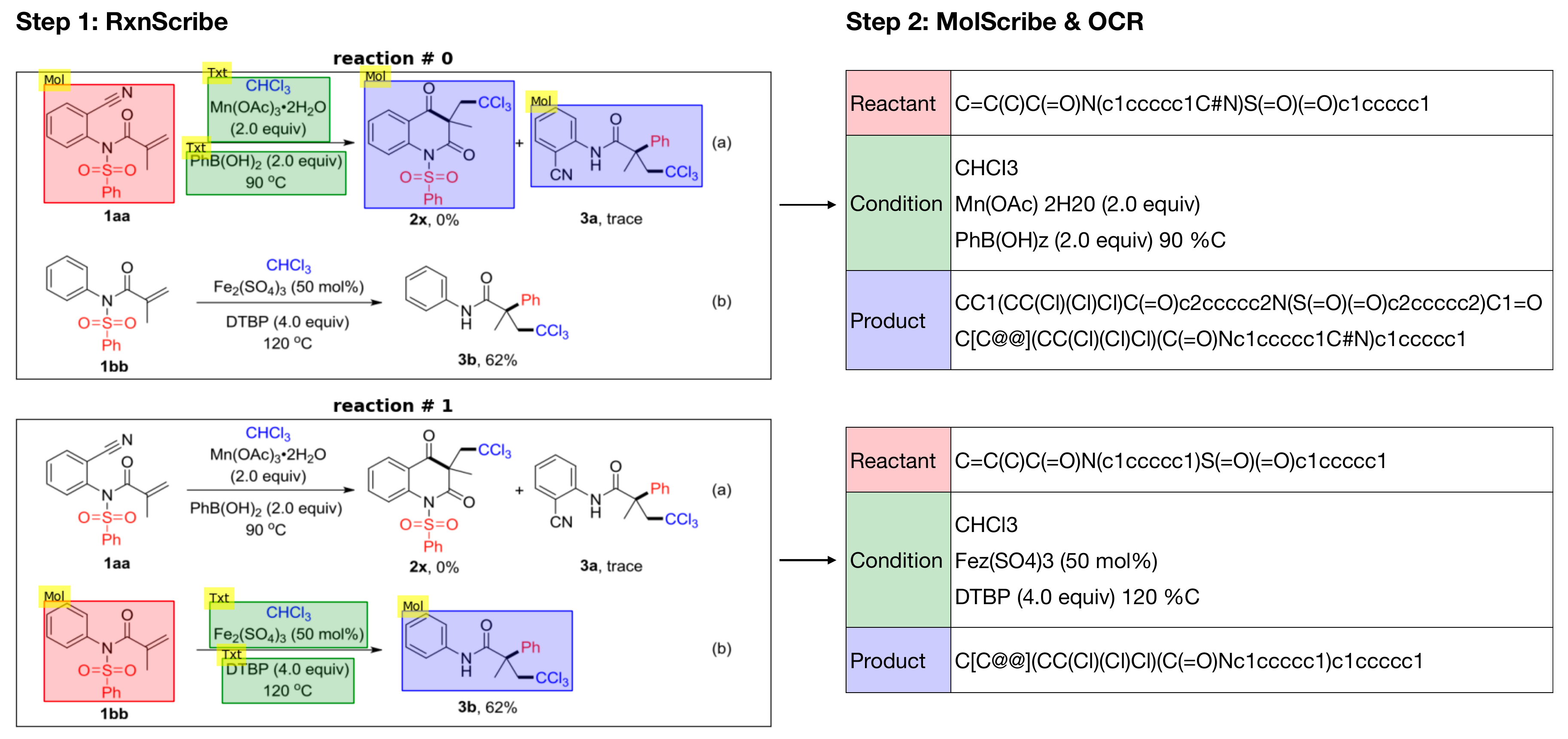}
    \caption{Illustration of extracting reactions from a figure. First, RxnScribe parses the two reactions in the figure. Then, the molecules and text are recognized by MolScribe and OCR models respectively. }
    \label{fig:rxn_from_figure}
\end{figure}

Reaction schemes are often defined graphically within figures and in a wide range of styles, requiring a sophisticated level of visual understanding to correctly extract. To this end, \ours incorporates RxnScribe \cite{RxnScribe}, a model previously designed for the extraction of reaction schemes from figures. \Cref{fig:rxn_from_figure} demonstrates the extraction process on a figure with two reactions. In it, RxnScribe predicts the structure of both reactions correctly, with reactants, conditions, and products highlighted in red, green, and blue boxes, respectively.

\paragraph{Molecule Recognition}
Molecule recognition is the task of translating an image of a molecule into its corresponding chemical structure, typically represented as a SMILES string in a computer-readable format. \ours includes MolScribe \cite{MolScribe}, a model we developed earlier for molecule recognition. Our previous modules for text-figure coreference and reaction diagram parsing only extract the high-level structure of diagrams. To fully extract the reaction or molecular information, we further crop the bounding boxes and pass the individual subdiagrams to downstream modules. As shown in \Cref{fig:mol_from_figure}, we use MolScribe and EasyOCR\cite{easyocr}, an off-the-shelf optical character recognition tool, to translate the content in each bounding box to paired SMILES strings and text labels. Similarly, in \Cref{fig:rxn_from_figure}, we use the same tools to extract the molecular structures of products and reactants, as well as text descriptions of accompanying reaction conditions. 
\section{Text Analysis}

\begin{figure}[t]
    \centering
    \includegraphics[width=\linewidth]{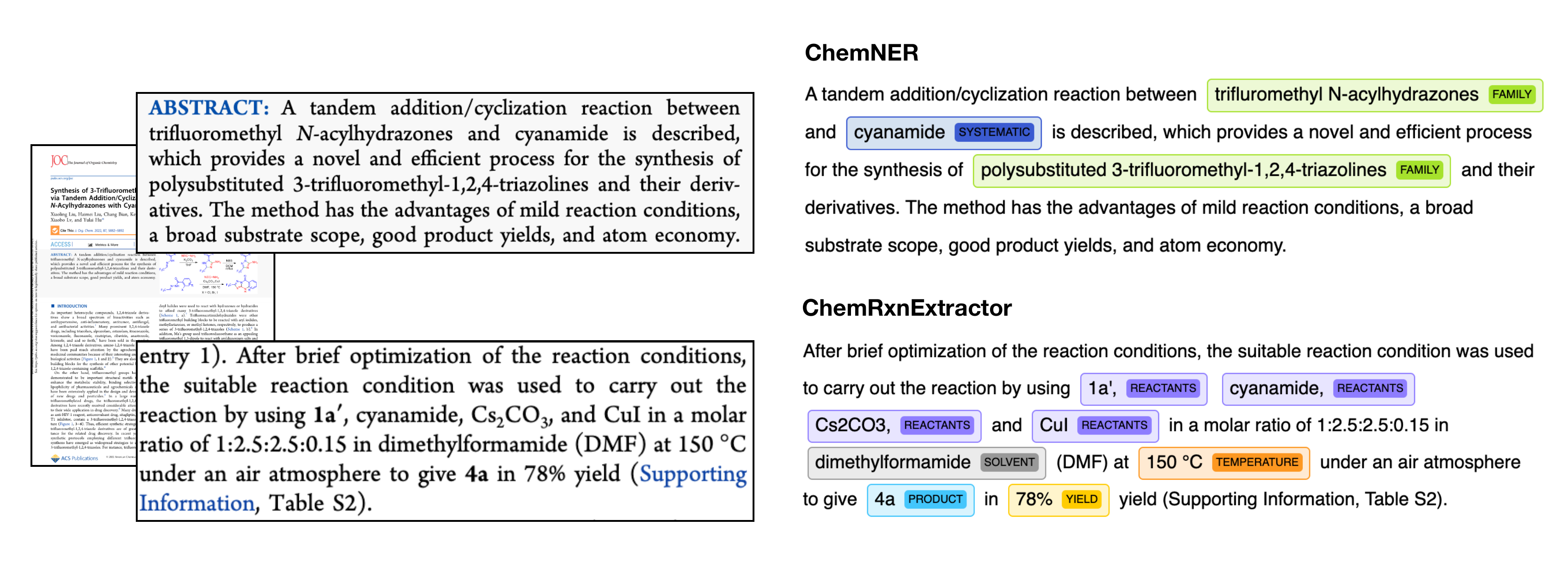}
    \caption{Illustration of extracting chemical entities and reactions from text. The passages are drawn from \citeauthor{doi:10.1021/acs.joc.2c00176}.\cite{doi:10.1021/acs.joc.2c00176}}
    \label{fig:rxn_from_text}
\end{figure}

 As seen in \Cref{fig:rxn_from_text}, \ours contains two powerful natural language processing models which excel at extracting chemical entities and reactions from the text in chemistry literature.able to extract chemical entities and reactions from text. The machine learning models are named ChemNER and ChenRxnExtractor respectively.

\paragraph{Named Entity Recognition} 
Our first model is dedicated to the task of extracting chemical entities from a given text excerpt. In scientific literature, chemical entities can take on diverse forms, including molecular formulas (e.g., NaOH), IUPAC systematic names (e.g., 1,3,4-oxadiazole), abbreviations (e.g., GABA), or database identifiers (e.g., CID16020046). For successful extraction, it is essential to locate these mentions in the text and accurately determine their specific forms.

In \ours, we provide a model named ChemNER, which is trained on the publicly available CHEMDNER corpus \cite{CHEMDNER}. This corpus comprises a collection of PubMed abstracts with expert-annotated chemical entity mentions. Our model adopts a sequence tagging approach using the BIO format. We fine-tune a language model that has been pre-trained on biomedical literature,\cite{biobert} further enhancing the model's performance and domain-specific understanding. In \Cref{fig:rxn_from_text}, ChemNER detects all three chemical entities and correctly makes the distinction that ``cyanamide" refers to a specific compound whereas the other chemical mentions refer to general families of compounds. 

\paragraph{Reaction Extraction}
Reaction extraction is a structured prediction task that involves identifying the reactions presented in the text. \ours includes ChemRxnExtractor \cite{chemrxnextractor}, a model previously developed for text-based reaction extraction. \Cref{fig:rxn_from_text} displays a processed reaction description where the product is represented by the identifier ``4a", with additional information about reaction conditions and yield also highlighted. However, the chemical structures of the extracted reactant ``1a" and product ``4a" are omitted in the text, highlighting the importance of our models for coreference resolution and molecule recognition in diagrams.
\section{Multimodal Integration}

Complete reaction schemes, which require full structural information of reactants and products as well as complex descriptions of reaction conditions, are often specified across multiple paragraphs, tables, and figures. Understanding the connections between these modalities is challenging, as demonstrated in \Cref{fig:lit example}, and has not been significantly explored by previous works. \ours begins to address the general task of multimodal integration by dividing the problem into two main challenges. For one, detailed reaction condition and yield data must be properly aligned with machine readable molecular structures of the reactions they refer to. Furthermore, many diagrams are underspecified, and the R-groups they contain must be inferred from a separate molecule in the diagram or a completely different table altogether. In the following sections, we describe how we integrate our individual model results together for multimodal understanding.

\paragraph{Reaction Condition Alignment}

\begin{figure}[t]
    \centering
    \includegraphics[width=\linewidth]{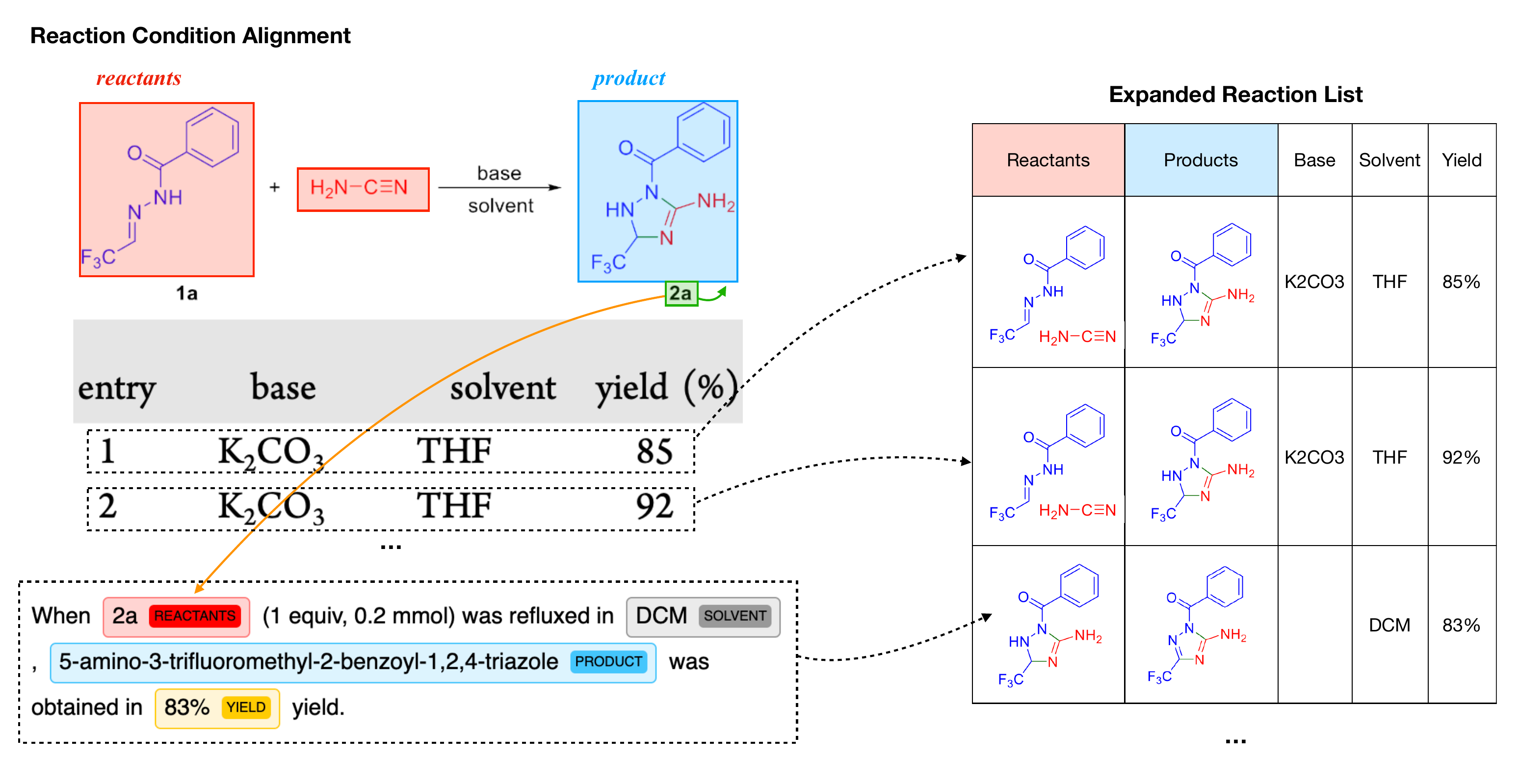}
    \caption{Reaction Condition Alignment. We augment incomplete reaction descriptions in text with resolved molecule identifier pairs and parse additional reaction condition tables. Example adapted from \citeauthor{doi:10.1021/acs.joc.2c00176}.\cite{doi:10.1021/acs.joc.2c00176}}
    \label{fig:rxn_condition_overview}
\end{figure}

In \ours, we provide methods to align the reaction data contained in figures with the information from text and tables, obtaining more complete reaction descriptions. 

One type of reaction condition alignment we address is the task of integrating information from condition screening tables with their corresponding reactions displayed in figures, such as in \Cref{fig:rxn_condition_overview}. For this, we created a parser to extract the table headers and columns. We use a dictionary-based classifier to categorize each column based on its header, such as being for temperature, solvent, yield, or other common types of metadata. Each row in the table corresponds to a complete configuration of reaction conditions, which we add to the set of reaction conditions for the relevant reaction. 

Reactions and their details are often described within the accompanying text as well. However, the reactants and products in this modality are often distinguished by their unique identifier, with the structural information of the molecules defined separately in figures. With only the identifier, these text-based reactions would be incomplete due to the missing of molecular structures. To address this issue, we align the molecular structure information from figures with their identifiers in text. From our figure analysis module, we first obtain a mapping between the identifiers and their structures using MolCoref. Whenever an identifier is encountered during the text-based reaction extraction stage, we substitute the identifiers with their SMILES representation. This integration along with our table-figure integration allows for the unification of information across three modalities to extract significantly more complete reaction data. 

\paragraph{R-Group Resolution}

\begin{figure}[t]
    \centering
    \includegraphics[width=\linewidth]{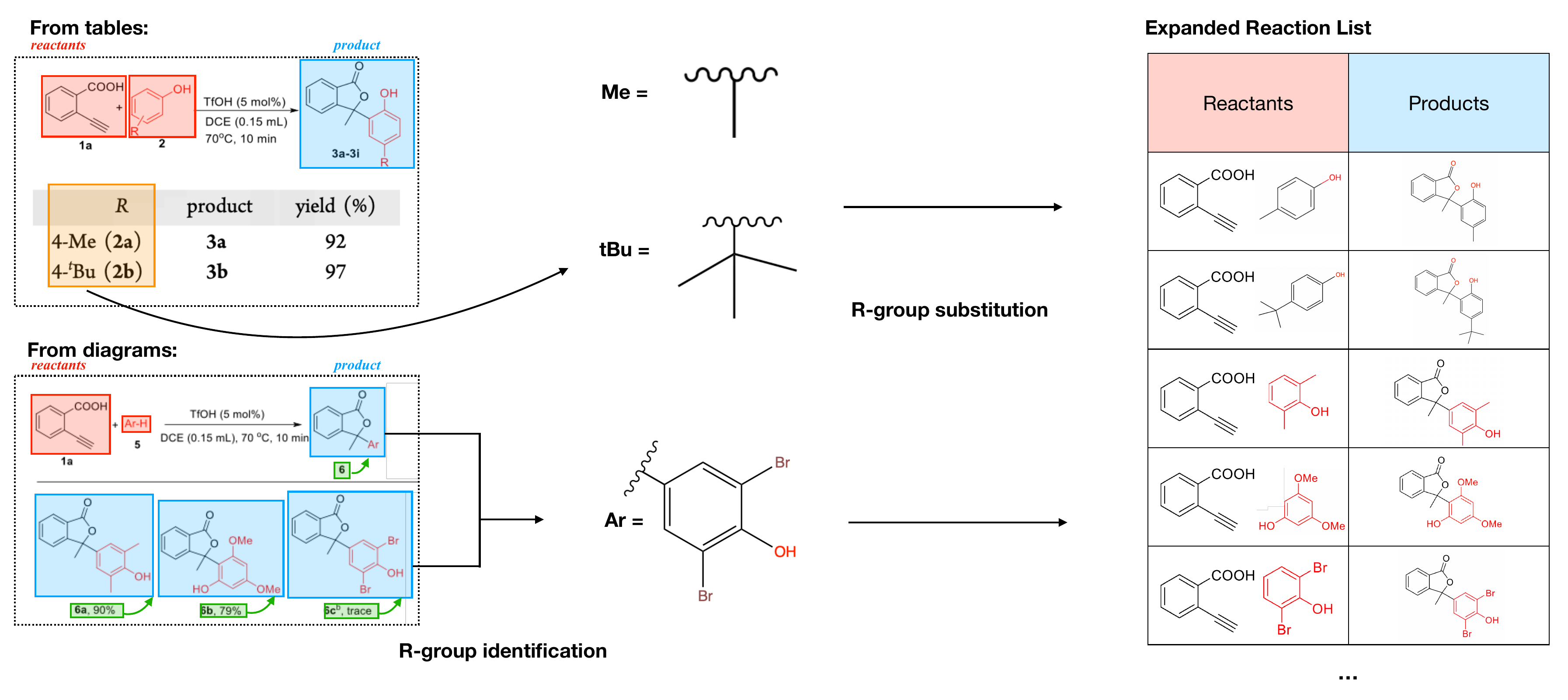}
    \caption{R-Group Resolution. R-groups are first identified from diagrams and tables and then substituted into the appropriate reactant molecule templates. Example adapted from Zhang et al.\cite{doi:10.1021/acs.joc.3c00760}}
    \label{fig:r_group_overview}
\end{figure}

Previous work from ChemSchematicResolver \cite{CSR} has been done to parse simple definitions of R-groups that are explicitly expressed as text chemical formulae within figures (e.g., ``R=Me") and perform the corresponding substitutions. In addition to this case, \ours seeks to comprehend other forms of substrate scope, namely the cases where products are depicted as different molecular structures or where a table separate from the reaction scheme defines the R-groups, which require further reasoning to determine the resolved molecular structures.

We address the two most common modes of presentation for substrate scope, which are shown in \Cref{fig:r_group_overview}. In the first case, the reaction template is displayed graphically and the R-groups are defined as text in an associated table or label. For this, we parse the R-group information from the text formula and use MolScribe\cite{MolScribe} to predict the graph structures of the template molecules. We then directly substitute the chemical formulas of the R-groups into their placeholders in the graphs. These structures are then expanded and converted to SMILES strings by MolScribe's postprocessing methods. 

In the second case, in addition to a reaction template, there is a set of possible products defined in the diagram with which one must infer the structures of the R-groups and the reactants. To approach this, we first leverage MolCoref to identify the labels of all molecules in the figure and match the label prefixes to associate the specific products with their template molecule. Given the reaction template and the specific product, we use a subgraph isomorphism algorithm implemented in RDKit \cite{rdkit2010} to identify the atom mapping between the two molecular structures. The unmapped atoms in the specific product are molecular fragments that correspond to the substructures of the R-groups. We substitute the identified R-group fragments into the reactant templates in order to obtain the full molecular structures of the entire reaction. 

With the two integration steps presented above, we obtain a reaction list with complete molecule structures and conditions. In the experiments, we evaluate the performance of the pipeline.
\section{Experiments}

One of the central challenges in developing a reaction extractor is a lack of high-quality benchmark datasets with corresponding evaluation metrics. To this end, we created our own dataset consisting of reactions and diagrams from chemistry literature, with manually produced annotations. In addition, we compared the system output against reactions in the Reaxys database. While our annotation scheme and extraction scope are not fully aligned with that of Reaxys (e.g., Reaxys does not include reactions with low or no yield), this challenging evaluation provides another measure of the system performance. We conduct a meticulous error analysis for both evaluation settings and further discuss the performance of individual modules in \ours. 

\paragraph{Evaluation With Annotated Data}

\begin{figure}[t]
    \centering
    \includegraphics[width = \linewidth]{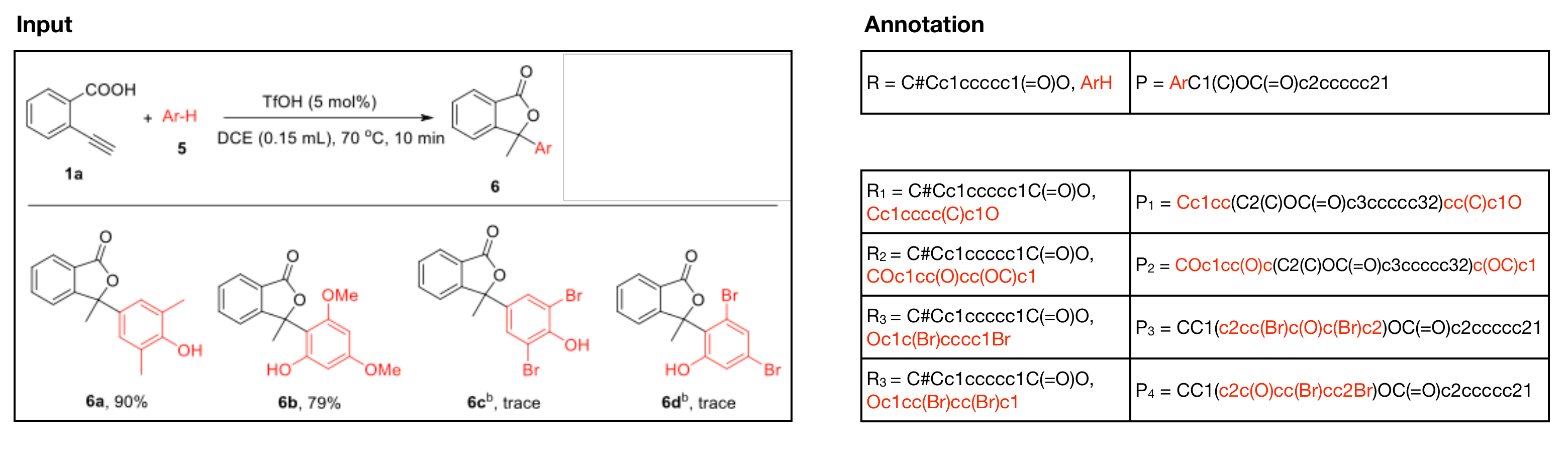}
    \caption{Illustration of annotation process, where we parse the SMILES strings of the template reaction $(R, P)$ and provide each detailed reaction $(R_i, P_i)$.}
    \label{fig:annotation}
\end{figure}

We evaluate \ours on a newly annotated reaction extraction dataset. This dataset contains 1007 reactions collected from 78 figures from recent issues of five chemistry journals: Journal of Organic Chemistry, Organic Letters, Angewandte Chemie International Edition, European Journal of Organic Chemistry, and Asian Journal of Organic Chemistry. The figures in this dataset are substrate scope diagrams. Using ChemDraw, we annotated SMILES strings for every reaction by inferring the structure of reactants from the structures of the template product and table of full products. A set of example annotations for this dataset is displayed in \Cref{fig:annotation}. For each substrate scope diagram, we first annotate the reaction template $(R, P)$, where $R$ and $P$ are the sets of SMILES strings for the reactants and products respectively, which may contain R-groups. Then, we annotate the substrate scope $\{(R_i, P_i)\}$, where $R_i$ is a set of reactants whose R-groups have been substituted based on $P_i$.

\begin{table*}[t]
\centering
\caption{Performance of \ours for extracting reactions from substrate scope diagrams, as well as the individual performance of each module in \ours.}
\label{tab:results}
\setlength{\tabcolsep}{3pt}
    \resizebox{0.8\linewidth}{!}{
\begin{threeparttable}
   \begin{tabular}{lc}
\toprule
Module & Evaluation Score\tnote{*}\\
\midrule
\ours & 79.1 / 62.0 / 69.5 \\
\midrule
\textit{Evaluation of individual models} \\
\ - Molecule Detection (MolDetect) & 86.0  \\
\ - Coreference Resolution (MolCoref) &  91.4 / 88.9 / 90.1\\
\ - Reaction Diagram Parsing (RxnScribe) & 91.9 / 90.1 / 91.0\\
\ - Molecule Recognition (MolScribe) & 71.9\\
\ - Named Entity Recognition (ChemNER) & 87.1 / 88.1 / 87.6\\
\ - Reaction Extraction (ChemRxnExtractor) &  79.3 / 78.1 / 78.7\\
\bottomrule
\end{tabular}
\begin{tablenotes}
\item[*] Precision/Recall/F1 by default. For molecule detection, we use Average Precision \cite{MSCOCO}. For molecular recognition, we use accuracy. 
\end{tablenotes}
\end{threeparttable}
}
\end{table*}

For this dataset, we evaluate the model's predictions using exact match, i.e., a predicted reaction $(\hat{R}, \hat{P})$ is considered correct only if all the molecular structures of its reactants and products match those in a ground truth reaction. We compute the precision, recall, and F1 to assess the model's performance. Here, the precision measures what fraction of the model's predictions is correct, the recall measures what fraction of the ground truth reactions is correctly predicted, and F1 score is the harmonic mean of precision and recall. As seen in \Cref{tab:results}, \ours achieves a precision of 79.1\%, recall of 62.0\% and F1 score of 69.5 \% on this task. 

Each individual module in \ours is also evaluated on an independent benchmark to measure its performance in isolation of the entire system. \Cref{tab:results} shows state-of-the-art performances of all six individual machine learning models on their respective benchmarks. In particular, MolCoref achieves an 90.1\% F1 score on a dataset of 1696 diagrams meticulously annotated with molecule-identifier information. As detailed in our past research, RxnScribe achieves a strong 91.0\% F1 score on identifying single line reaction diagrams, and MolScribe has an accuracy of 71.9\% on realistic molecular structures drawn from past ACS publications. We discuss additional evaluation details and error contribution rates to the next section.

\paragraph{Evaluation With Reaxys}

\begin{table}[t]
    \centering
    \begin{tabular}{l c c c }

    \toprule\\
     & Correct& Total Predictions& Accuracy \\
     \midrule 
     \ours & 257 & 400 & \textbf{64.3\%}\\
     %\ours (w/ Layout Parser) & 165 & 359 & 46.0 \\
      ReactionDataExtractor 2.0 & 9 & 102& 8.8\% \\
     \bottomrule
\end{tabular}
    \caption{Reaction extraction results on journal articles compared against Reaxys.}
    \label{tab:journal}
\end{table}

We evaluate the performance of \ours by comparing the extractions against those in Reaxys. Reaxys is a large commercial database of reactions that is periodically updated by chemical experts who manually extract the data from journal articles. 

We construct the dataset for this task by collecting 19 journal articles containing 155 figures from recent issues of The Journal of Organic Chemistry and Organic Letters that contained reaction condition and substrate scope screening tables. These journal articles were each converted into a triple of figures, texts, and tables for input to \ours with a set of off-the-shelf PDF-parsing tools corresponding to each modality\cite{layoutparser, pdfdataextractor, pdfminer}. Due to errors in diagram parsing frequently yielding inaccurate borders, we manually adjusted diagram segmentations for this dataset. Existing reaction extractions in Reaxys provided the groundtruth annotations. For each article, the groundtruth thus contains a list $\{(R_i, P_i)\}$ where $R_i$ and $P_i$ are sets of SMILES strings for the reactants and products respectively. 

We use a soft match to evaluate the accuracy of our pipeline's predictions. First, molecular structures are considered to be equivalent if they are tautomers to each other since some compounds rearrange to specific isomers in solution. Second, a predicted reaction $(\hat{R}, \hat{P})$ is considered correct if Reaxys contains an entry $(R, P)$ such that $\hat{R}$ is a subset of $R$ and $\hat{P}$ is a subset of $P$. We choose to use this evaluation metric because of ambiguities that arise during the annotation process, for example, whether certain compounds are considered reactants or reagents specified in the set of conditions instead. 

As seen in \cref{tab:journal} \ours extracts 400 reactions from this dataset, of which 257 have a soft match in the Reaxys database, for an accuracy of 64.3\%. Since ReactionDataExtractor 2.0\cite{reactiondataextractor2} does not extract from texts or tables, we only provide the segmented diagrams for each journal article to ReactionDataExtractor 2.0. It achieves an accuracy of 8.8\% with 102 total predictions in this evaluation setting. Besides reactions described in texts or tables, ReactionDataExtractor was also unable to resolve reactions whose depictions involved R-groups, which comprised the majority of reactions extracted by \ours. Moreover, we applied the same evaluation to the fully automatic version of \ours provided through our code package and web portal. In this setting, \ours achieved an accuracy of 46.0\% on 359 total predictions. The decrease in accuracy can mainly be attributed to inaccurate diagram segmentations during the PDF parsing process, since the automatic tool LayoutParser was not trained specifically on chemistry literature. Additional implementation details can be found in the Supporting Information. 

\paragraph{Analysis and Discussion}

\begin{figure}[t]
    \centering
    \includegraphics[width = 0.8\linewidth]{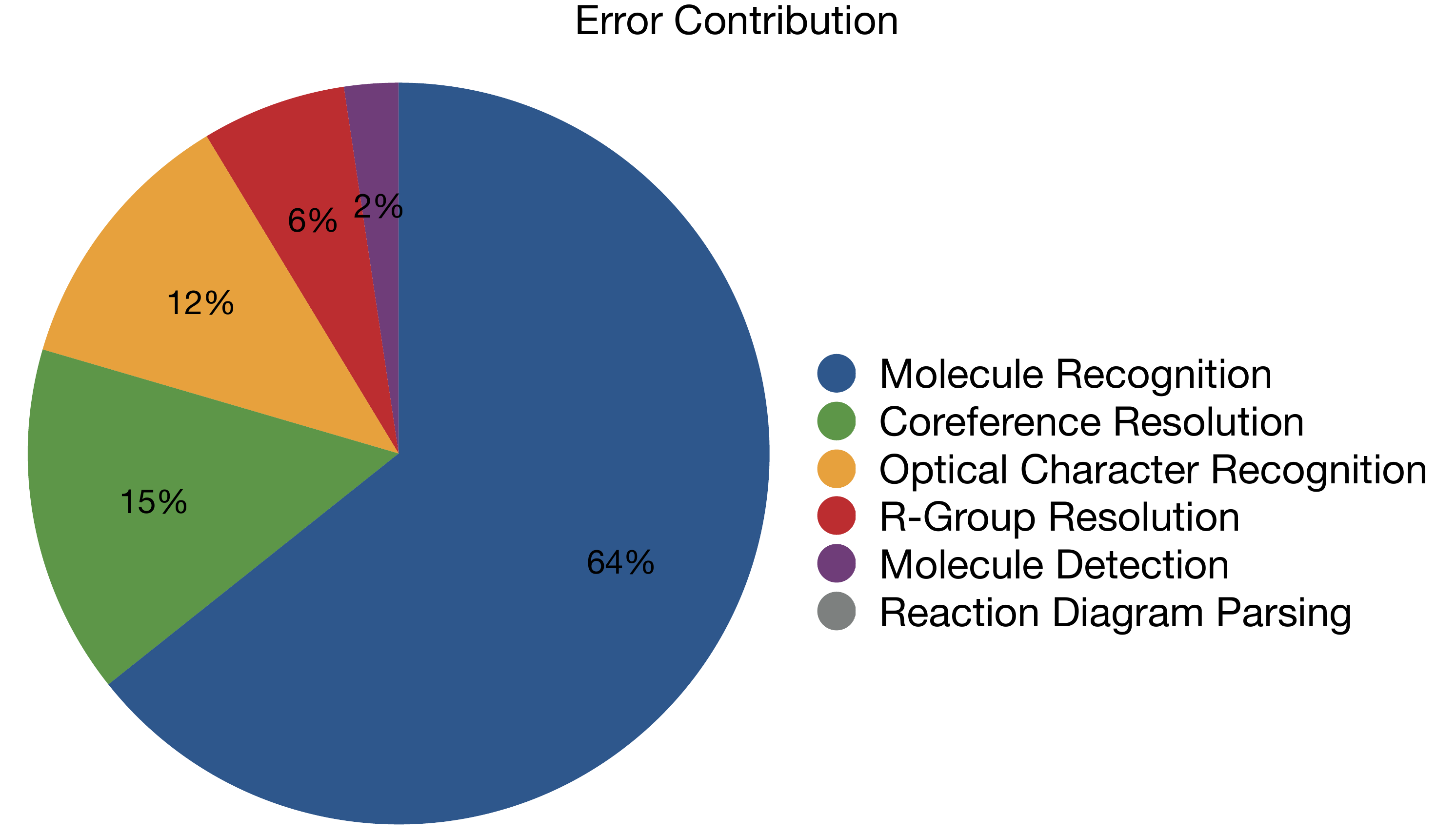}
    \caption{Error contribution of each relevant module to the R-group Resolution process}
    \label{fig:contribution}
\end{figure}

\begin{figure}
    \centering
    \includegraphics[width=0.95\linewidth]{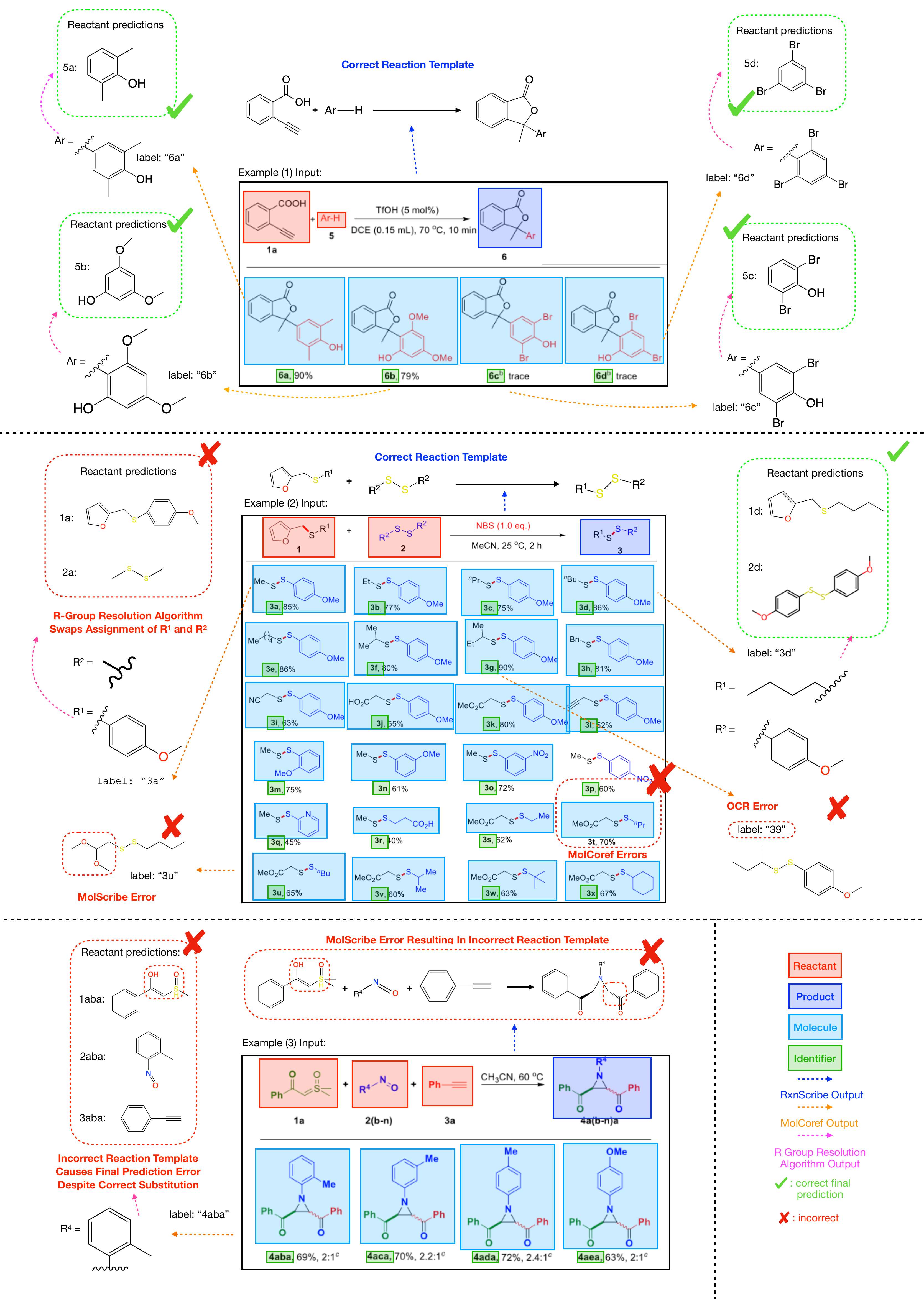}
   \caption{Examples of predictions and common errors of \ours on substrate scope diagrams.}
    \label{fig:r_group_sub}
\end{figure}

We analyze the error contribution of individual components in \ours. Since the majority of reactions extracted in the previous evaluations are described in substrate scope diagrams, we focus our analysis towards the performance on the task of extracting from this setting. \Cref{fig:contribution} displays the error contribution of each module in the R-group resolution process. The full evaluation scores of each model are displayed in \Cref{tab:results}. 

\Cref{fig:r_group_sub} illustrates examples of successful predictions and common errors by \ours. In Example (1), \ours is able to correctly identify the reaction template, determine the molecular bounding boxes, and resolve all of the coreferences correctly. The overall pipeline is able to extract all four reactions depicted in the substrate scope diagram correctly, which can be attributed to the strong performance of MolCoref and RxnScribe in the initial stage. We provide evaluation results of MolDetect and MolCoref in \Cref{tab:moldet_eval} and also compare the models against ChemSchematicResolver \cite{CSR}. MolDetect and MolCoref leverage the simple sequential learning framework to achieve strong performance in both tasks, whereas errors propagate throughout ChemSchematicResolver's rule-based pipeline. Per \Cref{fig:contribution}, MolCoref ultimately caused 15\% of incorrect predictions. This outsized error contribution was primarily due to diagrams in which one molecule had multiple labels, a presentation style not seen in the training dataset. On the other hand, RxnScribe achieves an F1 score of 91\% for parsing single line reaction diagrams, which make up the majority of reaction templates in substrate scope diagrams, and contributed to 0\% of overall errors. Our prior work\cite{RxnScribe} provides a more detailed quantitative evaluation of RxnScribe on extracting reactions from diagrams of various styles.

\begin{table}[t]
    \centering
    \begin{tabular}{l c c c c }

    \toprule
     & \multicolumn{1}{c}{Detection}  & \multicolumn{3}{c}{Coreference}\\
     & Average Precision& Precision& Recall & F1 \\
     \midrule 
     ChemSchematicResolver & 28.8 & 83.8 & 31.7&46.0\\
     MolDetect & $\mathbf{86.0}$ & --&-- & --\\
      MolCoref & $82.9$ & $\mathbf{91.4}$& $\mathbf{88.9}$& $\mathbf{90.1}$ \\
     \bottomrule
\end{tabular}
    \caption{Evaluation of chemical diagram entity detection and coreference performance (scores are in \%).}
    \label{tab:moldet_eval}
\end{table}

Our R-group resolution algorithm also performs robustly and is generally able to correctly identify the R-groups from each product and perform the corresponding substitutions in the reactant template when the input is free of errors. However, there were a small number of cases where the algorithm returned an incorrect prediction. For example, some product templates are completely symmetric but contain two different R-groups. Since the R-group resolution algorithm does not take into account information about the layout or color of the original diagram, it is unable to differentiate between the two correctly extracted R-group fragments. In other diagrams where errors occurred, specific presentation choices violated assumptions made in the design of the algorithm. Some authors switched the chirality of certain atoms between the template and product, and others included products where not every R-group in the original template had a substituent. A more detailed discussion of specific errors can be found in the Supporting Information. 

 In contrast, over half of the errors in the \ours pipeline occurred during molecule recognition. In \Cref{fig:r_group_sub}, example (2) displays a MolScribe error occurring on a MolCoref prediction, where a molecule with label \textbf{3u} is parsed incorrectly. Example (3) displays an instance where there is a MolScribe error in the original reaction template. From this, we observe that there are two reasons for MolScribe’s outsized error contribution. First, if there is a single MolScribe error in the original reaction template, the extraction results for the entire diagram will be incorrect. This scenario contributed 41.6\% of all errors. Second, MolScribe only achieves a 71.9\% accuracy on molecules from ACS publications, which are often drawn in diverse styles \cite{MolScribe}. Furthermore, the tool we employed for optical character recognition of molecule labels, EasyOCR, was another large source of error.  Many labels were parsed incorrectly. In example (2), the label ``\textbf{3g}" was mistakenly parsed as ``\textbf{39}", which meant that the product was not processed by the downstream algorithm, as it was not associated with the product with label ``\textbf{3}". 

\begin{table}[t]
    \centering
    \begin{tabular}{lcccc}

    \toprule
      % & \multicolumn{4}{c}{NER evaluation}\\
     & Percentage & Precision & Recall & F1 \\
     \midrule 
     ABBREVIATION &  16.5\%& 85.9 & 84.5&85.2  \\
     FAMILY  & 13.0\%&76.5& 85.4&  80.7  \\
     FORMULA  & 14.0\%&86.3& 84.8&  85.6 \\
     IDENTIFIER  & 2.4\%&92.2 & 78.1& 84.5   \\
     MULTIPLE  & 0.7\% & 68.7 & 74.4 &  71.5 \\
     SYSTEMATIC & 22.9\%  &89.8 & 88.6&  89.2  \\
     TRIVIAL  & 30.5\% &91.5& 93.5& 92.5  \\
     \midrule
     Overall  &100\%& 87.1 & 88.1 & 87.6   \\
     \bottomrule
    \end{tabular}
    \caption{Evaluation of chemical entity named entity recognition by entity type (scores are in \%).}
    \label{tab:ner_eval}
\end{table}

We further analyze the two text-based extraction models in \ours, namely ChemNER and ChemRxnExtractor. ChemNER is trained based on the BioBERT-large \cite{biobert} checkpoint using the CHEMDNER dataset.\cite{CHEMDNER} The dataset annotates mentions of chemical entities in seven types. Whereas past work does not make distinctions based on entity type during the evaluation, we evaluate the entity-level precision, recall and F1 scores on each type and also report the micro-averaged overall performance in \Cref{tab:ner_eval}. The model achieves an F1 score of 87.57 \% on the entire test set, with stronger performance on the two most common classes, SYSTEMATIC and TRIVIAL. Complete evaluation results for ChemRxnExtractor can be found in previous work \cite{chemrxnextractor}.
\section{\ours Interfaces}

\ours is accessible through two interfaces: (1) a comprehensive Python package that integrates all our models and utility functions, and (2) a user-friendly Web portal that simplifies the toolkit's usage, making it accessible to a wider audience, even those without programming knowledge.

\paragraph{Python Package}
We provide an open-source Python package (\url{https://github.com/CrystalEye42/OpenChemIE}) that integrates all our models and utility functions, including the PDF parser and the models for text, figure, and table analysis. We further implement methods that take a PDF document as input and effortlessly execute the information extraction pipeline, returning the extracted molecule and reaction data in a structured format. To ensure smooth usage, we provide detailed installation instructions and example use cases, enabling chemists with basic programming skills to efficiently process literature data using the toolkit. 

\begin{figure}[t]
    \centering
    \includegraphics[width=\linewidth]{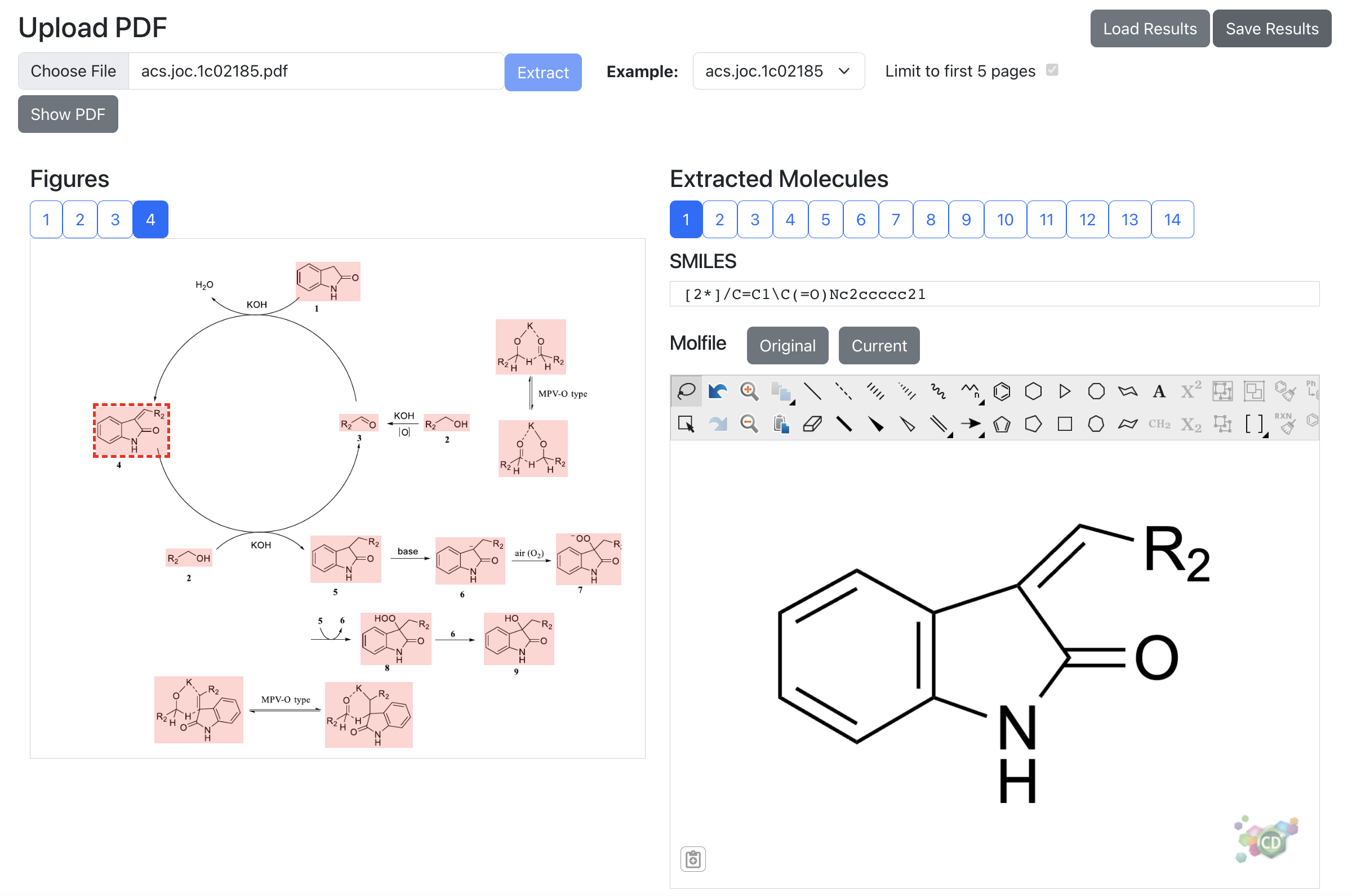}
    \caption{Illustration of the web interface for extracting molecular structures from PDF files. The uploaded PDF document is from \citeauthor{doi:10.1021/acs.joc.1c02185}.\cite{doi:10.1021/acs.joc.1c02185}}
    \label{fig:web interface}
\end{figure}

\paragraph{Web Portal}
We have developed a user-friendly web portal (\url{https://mit.openchemie.info}) that streamlines the PDF upload process, automatically executes extraction models, and conveniently displays the results. Users can upload a PDF document of a chemistry paper, which will be processed on our backend server using the \ours toolkit. The extraction results will be visualized on the portal. As in \Cref{fig:web interface}, predicted molecule structures are displayed in a web-based Ketcher editor, enabling the user to edit the model's predictions if desired. Due to computational constraints, our public portal can process a maximum of five pages from each paper. However, users can freely download and deploy the web portal on their own machines, granting access to all functionalities. 

\section{Conclusion}

In this paper we present \ours, a comprehensive system for information extraction from chemistry literature at the document level. \ours addresses the need for the integration of information across multiple modalities in order to provide complete extractions of molecules and reactions. We approach the general challenge of chemistry information extraction by incorporating chemistry-informed algorithms to integrate the results from individual modalities to obtain the final outputs. This approach allows for the extraction of previously unresolvable information, such as substrate scope investigations, and is a notable step toward achieving multimodal analysis of chemistry literature. 

\ours has made remarkable progress toward its objective of extracting reaction data comprehensively from chemical literature, although some challenges remain to be addressed. For instance, there is room for enhancing the performance of machine learning models on diverse literature data: MolScribe might be further developed to more precisely capture less common representations of molecular structures, including Markush structures; the PDF parsing tool may benefit from adjustments to better cater to chemical documents. Additionally, while our system is adept at parsing multiple multimodal relationships, enhancing its ability to understand the complex interdependencies between different modalities in chemical documents represents an exciting area for future development. The emergent abilities of large language models hold promise for providing a more integrated end-to-end solution for chemical information extraction, suggesting an optimistic pathway forward.

\section{Data and Software Availability}
The \ours toolkit is publicly available:
\begin{itemize}
    \item Source code: \url{https://github.com/CrystalEye42/OpenChemIE}
    \item Web interface: \url{https://mit.openchemie.info}
\end{itemize}. 
Individual machine learning models in \ours can be found at the following links:
\begin{itemize}
    \item MolScribe: \url{https://github.com/thomas0809/MolScribe}
    \item RxnScribe: \url{https://github.com/thomas0809/RxnScribe}
    \item MolDetect/MolCoref: \url{https://github.com/Ozymandias314/MolDetect}
    \item ChemNER: \url{https://github.com/Ozymandias314/ChemIENER}
    \item ChemRxnExtractor: \url{https://github.com/jiangfeng1124/ChemRxnExtractor}
\end{itemize}
The datasets for our molecule detection, molecule coreference resolution, and R-group resolution processes are constructed from journal articles shared between the American Chemical Society (ACS) and MIT under a private access agreement.
\begin{itemize}
    \item The annotated images for the molecule coreference and detection task, as well as their train/validation/test splits can be downloaded at \url{https://huggingface.co/datasets/Ozymandias314/MolCorefData}.
    \item  The diagrams and annotations for the R-group resolution dataset, as well as data for the comparison against Reaxys are located at \url{https://huggingface.co/datasets/Ozymandias314/OpenChemIEData} for download.
\end{itemize}

\begin{suppinfo}
Detailed evaluation results for our new models (ChemNER, MolDetect, and MolCoref) and the overall \ours pipeline, a description of the data annotation process, and implementation details for our PDF Parser are available in the supporting information. 

\end{suppinfo}

\begin{acknowledgement}
The authors thank Guy Zylberberg for his contribution to the web interface development, and Zhengkai Tu for his chemical expertise. The authors additionally thank the members of Regina Barzilay's group and Connor Coley's group for helpful discussion and feedback. This work was supported by the DARPA Accelerated Molecular Discovery (AMD) program under contract HR00111920025 and the Machine Learning for Pharmaceutical Discovery and Synthesis Consortium (MLPDS).

\end{acknowledgement}

\bibliography{reference}

%\appendix
%\input{section/appendix}

\end{document}